%% file: acl2019.tex
\documentclass[11pt,a4paper]{article}
\usepackage[hyperref]{acl2019}
\usepackage{times}
\usepackage{latexsym}

\usepackage{url}
\usepackage[normalem]{ulem}
\usepackage{multirow}
\usepackage{paralist}
\usepackage{adjustbox}
\usepackage{graphicx}
\usepackage[scaled=.90]{helvet}
\usepackage{amsfonts}
\usepackage{float}
\usepackage{cleveref}
\crefformat{section}{\S#2#1#3}
\crefformat{subsection}{\S#2#1#3}
\restylefloat{table}
\renewcommand\vec[1]{\overrightarrow{#1}}
\newcommand\cev[1]{\overleftarrow{#1}}

\aclfinalcopy % Uncomment this line for the final submission
\def\aclpaperid{832} %  Enter the acl Paper ID here

\newcommand\blfootnote[1]{% 
\begingroup 
\renewcommand\thefootnote{*}\footnotetext{#1}% 
\addtocounter{footnote}{0}% 
\endgroup 
}

\newcommand\bifootnote[1]{% 
\begingroup 
\renewcommand\thefootnote{$\dag$}\footnotetext{#1}% 
\addtocounter{footnote}{0}% 
\endgroup 
}

\title{Topic-Aware Neural Keyphrase Generation for Social Media Language}

\author{Yue Wang$^{1*}$~Jing Li$^{2\dag}$~Hou Pong Chan$^{1}$~Irwin King$^{1}$~Michael R. Lyu$^{1}$~Shuming Shi$^{2}$\\
$^{1}$Department of Computer Science and Engineering \\
The Chinese University of Hong Kong, HKSAR, China \\
$^{2}$Tencent AI Lab, Shenzhen, China\\
$^{1}$\texttt{\{yuewang, hpchan, king,lyu\}@cse.cuhk.edu.hk}\\
$^{2}$\texttt{\{ameliajli,shumingshi\}@tencent.com}\\}

\date{}

\begin{document}
\maketitle
\blfootnote{This work was partially done when Yue Wang was an intern at Tencent AI Lab.}
\bifootnote{Jing Li is the corresponding author.}
\input{sections/abstract.tex}

\input{sections/introduction}
\input{sections/related-work}
\input{sections/model}

\input{sections/exp-setup}

\input{sections/exp-result}

\input{sections/conclusion}
\input{sections/ack}

% The acknowledgments should go immediately before the references.  Do
% not number the acknowledgments section. Do not include this section
% when submitting your paper for review. \\

% \noindent \textbf{Preparing References:} \\
% Include your own bib file like this:
% \verb|\bibliographystyle{acl_natbib}|
% \verb|\bibliography{acl2019}| 

% where \verb|acl2019| corresponds to a acl2019.bib file.
\bibliography{acl2019}
\bibliographystyle{acl_natbib}

% \appendix

% \section{Appendices}
% \label{sec:appendix}

% \verb|\appendix| before any appendix section to switch the section
% numbering over to letters.

% \section{Supplemental Material}
% \label{sec:supplemental}

\end{document}

%% file: sections/abstract.tex
\begin{abstract}
A huge volume of user-generated content is daily produced on social media. 
To facilitate automatic language understanding, we study keyphrase prediction, distilling salient information from massive posts.
While most existing methods \emph{extract} words from source posts to form keyphrases, we propose a sequence-to-sequence (seq2seq) based neural \emph{keyphrase generation} framework, enabling absent keyphrases to be created.
Moreover, our model, being \emph{topic-aware}, allows joint modeling of corpus-level latent topic representations, which helps alleviate the data sparsity that widely exhibited in social media language.
Experiments on three datasets collected from English and Chinese social media platforms show that our model significantly outperforms both extraction and generation models that do not exploit latent topics.\footnote{Our data and code are publicly released in \url{https://github.com/yuewang-cuhk/TAKG}} 
Further discussions show that our model learns meaningful topics, which interprets its superiority in social media keyphrase generation.
\end{abstract}
% \footnote{To obtain our datasets, please contact Yue Wang.}

%% file: sections/introduction.tex
\section{Introduction}

As social media continues its worldwide expansion, the last decade has witnessed the revolution of interpersonal communication.
While empowering individuals with richer and fresher information, the flourish of social media also results in millions of posts generated on a daily basis.
Facing a sheer quantity of texts, language understanding has become a daunting task for human beings.
Under this circumstance, there exists a pressing need for developing automatic systems capable of absorbing massive social media texts and figuring out what is important.

\input{tables/intro_example.tex}

In this work, we study the prediction of \textbf{keyphrases}, generally formed with words or phrases reflecting main topics conveyed in input texts~\cite{DBLP:conf/naacl/ZhangLSZ18}.
Particularly, we focus on producing keyphrases for social media language, proven to be beneficial to a broad range of applications, such as instant detection of trending events~\cite{weng2011event}, summarizing public opinions~\cite{DBLP:conf/kdd/MengWLZLW12}, analyzing social behavior~\cite{ruths2014social}, and so forth. 

In spite of the substantial efforts made in social media keyphrase identification, most progress to date has focused on \emph{extracting} words or phrases from source posts, thus failing to yield keyphrases containing absent words (i.e., words do not appear in the post).
Such cases are indeed prominent on social media, mostly attributed to the informal writing styles of users therein.
For example, Table~\ref{tables:intro-example} shows a tweet $S$ tagged with keyphrase ``\textit{super bowl}'' by its author, though neither ``\textit{super}'' nor  ``\textit{bowl}'' appears in it.\footnote{Following common practice~\cite{DBLP:conf/emnlp/ZhangWGH16,DBLP:conf/naacl/ZhangLSZ18}, we consider author-annotated hashtags as tweets' keyphrases.}
In our work, distinguishing from previous studies, we approach social media keyphrase prediction with a \emph{sequence generation} framework, which is able to create absent keyphrases beyond source posts.

Our work is built on the success of deep keyphrase generation models based on neural sequence-to-sequence (seq2seq) framework~\cite{DBLP:conf/acl/MengZHHBC17}.
However, existing models, though effective on well-edited documents (e.g., scientific articles), will inevitably encounter the data sparsity issue when adapted to social media.
It is essentially due to the informal and colloquial nature of social media language, which results in limited features available in the noisy data.
For instance, only given the words in $S$ (Table \ref{tables:intro-example}), it is difficult to figure out why ``\textit{super bowl}'' is its keyphrase. 
However, by looking at tweets $T_1$ to $T_3$, we can see ``\textit{yellow pants}'' is relevant to ``\textit{steelers}'', a \textit{super bowl} team. 
As ``\textit{yellow}'' and ``\textit{pants}'' widely appear in tweets tagged with ``\textit{super bowl}', it becomes possible to identify  ``\textit{super bowl}'' as $S$'s keyphrase.

Here we propose a novel \emph{topic-aware neural keyphrase generation model} that leverages latent topics to enrich useful features.
Our model is able to identify topic words, naturally indicative of keyphrases, via exploring post-level word co-occurrence patterns, such as ``\textit{yellow}'' and ``\textit{pants}'' in $S$.   
Previous work have shown that corpus-level latent topics
can effectively alleviate data sparsity in other tasks~\cite{DBLP:conf/emnlp/ZengLSGLK18, DBLP:journals/coling/LiSWW18}.
The effects of latent topics, nevertheless, have never been explored in existing keyphrase generation research, particularly in the social media domain.
To the best of our knowledge, \emph{our work is the first to study the benefit of leveraging latent topics on social media keyphrase generation}.
Also, our model, taking advantage of the recent advance of neural topic models~\cite{DBLP:conf/icml/MiaoGB17}, enables end-to-end training of latent topic modeling and keyphrase generation.

We experiment on three newly constructed social media datasets. 
Two are from English platform Twitter and StackExchange, and the other from Chinese microblog Weibo.
The comparison results over both extraction and generation methods show that our model can better produce keyphrases, significantly outperforming all the comparison models without exploiting latent topics.
For example, on Weibo dataset, our model achieves $34.99\%$ F1@1 compared with $32.01\%$ yielded by a state-of-the-art keyphrase generation model~\cite{DBLP:conf/acl/MengZHHBC17}.
We also probe into our outputs and find that meaningful latent topics can be learned, which can usefully indicate keyphrases.  
At last, a preliminary study on scientific articles shows that latent topics work better on text genres with informal language style.

%% file: tables/intro_example.tex
\begin{table}[t]
\centering
\scalebox{0.92}{
\begin{tabular}{|p{7.8cm}|}
\hline
\underline{\textbf{Source post with keyphrase ``\textit{super bowl}'':}}\\
$[S]$: Somewhere, a wife that is not paying attention to the \textcolor{blue}{\textit{game}}, says "I want the \textcolor{blue}{\textit{team}} in \textcolor{blue}{\textit{yellow pants}} to \textcolor{blue}{\textit{win}}." \\
\hline
\hline
\underline{\textbf{Relevant tweets}:}\\

$[T_1]$: I been a \textcolor{blue}{\textit{steelers fan}} way before \textcolor{blue}{\textit{black}} \& \textcolor{blue}{\textit{yellow}} and this \textcolor{blue}{\textit{super bowl}}!\\
$[T_2]$: I will bet you the \textcolor{blue}{\textit{team}} with \textcolor{blue}{\textit{yellow pants wins}}.\\
$[T_3]$: Wiz Khalifa song '\textcolor{blue}{\textit{black}} and \textcolor{blue}{\textit{yellow}}" to spur the \textcolor{blue}{\textit{pittsburgh steelers}} and Lil Wayne is to sing "\textcolor{blue}{\textit{green}} and \textcolor{blue}{\textit{yellow}}' for the \textcolor{blue}{\textit{packers}}. \\
\hline
\end{tabular}
}
\vskip -0.5em
\caption{Sample tweets tagged with ``\textit{super bowl}'' as their keyphrases. \textcolor{blue}{\textit{Blue and italic words}} can indicate the topic of super bowl.
}\label{tables:intro-example}
\end{table}

%% file: sections/related-work.tex
\section{Related Work}

Our work is mainly in the line of two areas: keyphrase prediction and topic modeling. We introduce them in turn below.

\paragraph{Keyphrase Prediction.} 
Most previous efforts on this task adopt supervised or unsupervised approaches based on \emph{extraction} --- words or phrases selected from source documents to form keyphrases.
% Most of the methods are either supervised or unsupervised. 
Supervised methods are mostly based on sequence tagging~\cite{DBLP:conf/emnlp/ZhangWGH16,DBLP:conf/aaai/GollapalliLY17} or binary classification using various features~\cite{DBLP:conf/dl/WittenPFGN99,DBLP:conf/emnlp/MedelyanFW09}.
For unsupervised methods, they are built on diverse algorithms, including graph ranking~\cite{mihalcea2004textrank,DBLP:conf/aaai/WanX08}, document clustering~\cite{DBLP:conf/emnlp/LiuLZS09, DBLP:conf/emnlp/LiuHZS10}, and statistical models like TF-IDF~\cite{salton1986introduction}. 

Our work is especially in the line of social media keyphrase prediction, where extractive approaches are widely employed~\cite{DBLP:conf/emnlp/ZhangWGH16,DBLP:conf/naacl/ZhangLSZ18}.
On the contrary, we predict keyphrases in a \emph{sequence generation} manner, allowing the creation of absent keyphrases.
Our work is inspired by seq2seq-based keyphrase generation models~\cite{DBLP:conf/acl/MengZHHBC17,DBLP:conf/emnlp/ChenZ0YL18,chen2019integrated_dkg,DBLP:conf/aaai/wchen19}, which are originally designed for scientific articles.
However, their performance will be inevitably compromised when directly applied to social media language owing to the data sparsity problem.
Recently, \citet{DBLP:conf/naacl/yuewang19} propose a microblog hashtag generation framework, which explicitly enriches context with user responses.  
%specifically on hashtag generation that proposes to enrich the contexts by incorporating the user comments.
Different from them, we propose to leverage corpus-level latent topic representations, which can be learned without requiring external data. %data augmentation,
Its potential usefulness on keyphrase generation has been ignored in previous research and will be extensively studied here.

\paragraph{Topic Modeling.} 
Our work is closely related with topic models that discover latent topics from word co-occurrence in document level.
They are commonly in the fashion of latent Dirichlet allocation (LDA) based on Bayesian graphical models \cite{DBLP:journals/jmlr/BleiNJ03}.
These models, however, rely on the expertise involvement to customize model inference algorithms.  
Our framework exploits the recently proposed neural topic models~\cite{DBLP:conf/icml/MiaoGB17,srivastava2017autoencoding} to infer latent topics, which facilitate end-to-end training with other neural models and do not require model-specific derivation. 
It has proven useful for citation recommendation~\cite{DBLP:conf/cikm/BaiCLKX18} and conversation understanding~\cite{DBLP:journal/tacl/ZengLHGLK19}.
In particular, \citet{DBLP:conf/emnlp/ZengLSGLK18} propose to jointly train topic models and short text classification, which cannot fit our scenario due to the large diversity of the keyphrases~\cite{DBLP:conf/naacl/yuewang19}.
Different from them, our latent topics are learned together with language generation, whose effects on keyphrase generation have never been explored before in existing work.
%Also, previous work~\cite{DBLP:conf/naacl/yuewang19} have shown that language generation framework can better fit our keyphrase annotation task.
% \textcolor{cyan}{Also, previous work have shown that language generation can better fit .}
%It is worthy to note that the classification-based model proposed by~\citet{DBLP:conf/emnlp/ZengLSGLK18} cannot fit our scenario due to the large diversity of the keyphrases.
%However, none of the existing work explores the effects of latent topics on keyphrase generation, which is a gap our work fills in.

%% file: sections/model.tex
\section{Topic-Aware Neural Keyphrase Generation Model}\label{sec:model}
\input{figure_input/framework.tex}

In this section, we describe our framework that leverages latent topics in neural keyphrase generation. Figure~\ref{fig:framework} shows our overall architecture consisting of two modules --- a neural topic model for exploring latent topics (\cref{ssec:ntm_model}) and a seq2seq-based model for keyphrase generation (\cref{ssec:keyphrase_model}). 

% Before starting with more details, we first introduce the formulations of inputs.
Formally, given a collection $\mathcal{C} $ with $|C|$ social media posts $\{{\bf x}_1, {\bf x}_2, ..., {\bf x}_{|C|}\}$ as input, we process each post $\bf x$ into bag-of-words (BoW) term vector ${\bf x}_{bow}$ and word index sequence vector ${\bf x}_{seq}$.
${\bf x}_{bow}$ is a $V$-dim vector over the vocabulary ($V$ being the vocabulary size).
It is fed into the neural topic model following the BoW assumption~\cite{DBLP:conf/icml/MiaoGB17}.
${\bf x}_{seq}$ serves as the input for the seq2seq-based keyphrase generation model.

Below we first introduce our two modules and then describe how they are jointly trained (\cref{ssec:joint_learn}).

\subsection{Neural Topic Model}\label{ssec:ntm_model}

Our neural topic model (NTM) module is inspired by \citet{DBLP:conf/icml/MiaoGB17} based on variational auto-encoder~\cite{kingma2013auto}, which consists of an encoder and a decoder to resemble the data reconstruction process.

Specifically, the input ${\bf x}_{bow}$ is first encoded into a continuous latent variable $\bf z$ (representing $\bf x$'s topic) by a BoW encoder. 
Then the BoW decoder, conditioned on $\bf z$, attempts to reconstruct $\bf x$ and outputs a BoW vector ${\bf x'}_{bow}$.
Particularly, the decoder simulates topic model's generation process. We then describe their division of labor.

\paragraph{BoW Encoder.} 
The BoW encoder is responsible for estimating prior variables $\mu$ and $\sigma$, which will be used to induce intermediate topic representation $\bf z$. 
We adopt the following formula:
\begin{equation}
    \mu=f_\mu (f_e({\bf x}_{bow})),\, log\, \sigma =f_\sigma (f_e({\bf x}_{bow})),
\end{equation}
where $f_*(\cdot)$ is a neural perceptron with an ReLU-activated function following~\citet{DBLP:conf/emnlp/ZengLSGLK18}. 

\paragraph{BoW Decoder.} Analogous to LDA-style topic models, it is assumed that there are $K$ topics underlying the given corpus $\mathcal{C}$. 
Each topic $k$ is represented with a topic-word distribution $\phi_k$ over the vocabulary, and each post ${\bf x} \in \mathcal{C}$ has a topic mixture denoted by $\theta$, a $K$-dim distributional vector.
Specifically in neural topic model, $\theta$ is constructed by Gaussian softmax  \cite{DBLP:conf/icml/MiaoGB17}.
The decoder hence takes the following steps to simulate how each post ${\bf x}$ is generated:
\begin{compactitem}
\item Draw latent topic variable $\bf z \sim \mathcal{N}(\mu, \sigma^2)$
\item Topic mixture $\theta = softmax(f_\theta({\bf z}))$
\item For each word $w\in {\bf x}$
\begin{compactitem}
\item Draw $w\sim softmax(f_\phi(\theta))$
\end{compactitem}
\end{compactitem}
Here $f_*(\cdot)$ is also a ReLU-activated neural perceptron for inputs. 
In particular, we employ the weight matrix of $f_\phi (\cdot)$ as the topic-word distributions ($\phi_1, \phi_2, ..., \phi_K$).
In the following, we adopt the topic mixture $\theta$ as the topic representations to guide keyphrase generation.

\subsection{Neural Keyphrase Generation Model}\label{ssec:keyphrase_model}

Here we describe how we generate keyphrases with a topic-aware seq2seq model, which incorporates latent topics (learned by NTM) in its generation process. Below comes more details.

\paragraph{Overview.} 
The keyphrase generation module (KG model) is fed with source post $\bf x$ in its word sequence form ${\bf x}_{seq}=\langle w_1, w_2, ..., w_{|{\bf x}|}\rangle$ ($|{\bf x}|$ is the number of words in $\bf x$).
Its target is to output 
a word sequence $\bf y$ as $\bf x$'s keyphrase.
Particularly, for a source post with multiple gold-standard keyphrases, we follow the practice in~\citet{DBLP:conf/acl/MengZHHBC17} to pair its copies with each of the gold standards to form a training instance. 

To generate keyphrases for source posts, the KG model employs a seq2seq model.
The \textbf{sequence encoder} distills indicative features from an input source post.
The decoder then generates its keyphrase, conditioned on the encoded features and the latent topics yielded by NTM (henceforth \textbf{topic-aware sequence decoder}).

\paragraph{Sequence Encoder.}

We employ a bidirectional gated recurrent unit (Bi-GRU)~\cite{DBLP:conf/emnlp/ChoMGBBSB14} to encode the input source sequence.
Each word $w_i\in {\bf x}_{seq}$ ($i=1,2,...,|{\bf x}|$) is first embedded into an embedding vector $\mathbf{\nu}_i$, and then mapped into forward and backward hidden states (denoted as $\vec{{\bf h}_i}$ and $\cev{{\bf h}_i}$) with the following defined operations:
\begin{equation}
    \vec{{\bf h}_i} = f_{GRU}(\mathbf{\nu}_i, {\bf h}_{i-1}), 
\end{equation}
\begin{equation}
    \cev{{\bf h}_i} = f_{GRU}(\mathbf{\nu}_i, {\bf h}_{i+1}).
\end{equation}
The concatenation of $\vec{{\bf h}_i}$ and $\cev{{\bf h}_i}$, $[\vec{{\bf h}_i};\cev{{\bf h}_i}]$, serves as $w_i$'s hidden state in encoder, denoted as ${\bf h}_i$. Finally, we construct a memory bank: ${\bf M}=\langle{\bf h}_1, {\bf h}_2, ..., {\bf h}_{|{\bf x}|}\rangle$, for decoder's attentive retrieval.

\paragraph{Topic-Aware Sequence Decoder.}
In general, conditioned on the memory bank $\bf M$ and latent topic $\theta$ from NTM, we define the process to generate its keyphrase $\bf y$ with the following probability:
\begin{equation}
    Pr({\bf y}\,|\,{\bf x}) = \prod_{j=1}^{|\bf y|} Pr(y_j\,|\,{\bf y}_{<j}, \mathbf{M}, \theta),
\end{equation}
where ${\bf y}_{<j}=\langle y_1, y_2, ..., y_{j-1}\rangle$. 
And $Pr(y_j|{\bf y}_{<j},\mathbf{M},\theta)$, denoted as $p_j$, is a word distribution over vocabulary, reflecting how likely a word to fill in the $j$-th slot in target keyphrase. Below we describe the procedure to obtain $p_j$.

Our sequence decoder employs a unidirectional GRU layer.
Apart from the general state update, the $j$-th hidden state $\mathbf{s}_j$ is further designed to 
take input $\bf x$'s topic mixture $\theta$ into consideration:
\begin{equation}\label{eq:dec_input}
    \mathbf{s}_j =  f_{GRU}( [\mathbf{u}_j; \theta], \mathbf{s}_{j-1}),
\end{equation}
where $\mathbf{u}_j$ is the $j$-th embedded decoder input\footnote{We take the previous word from gold standards in training by teacher forcing and from the predicted word in test.} and $\mathbf{s}_{j-1}$ is the previous hidden state. Here $[;]$ denotes the concatenation operation.

The decoder also looks at $\mathbf{M}$ (learned by sequence encoder) and puts an attention on it to capture important information. 
When predicting the $j$-th word in keyphrase, the attention weights on $w_i\in {\bf x}_{seq}$ is defined as:
    \begin{equation}
    \alpha_{ij}=\frac{\exp(f_{\alpha}(\mathbf{h}_i, \mathbf{s}_j, \theta))}
    {\sum_{i'=1}^{|\mathbf{x}|} \exp(f_{\alpha}(\mathbf{h}_{i'},\mathbf{s}_j,\theta))},
    \end{equation}
where
    \begin{equation}\label{eq:attn_score}
     f_{\alpha}(\mathbf{h}_i,\mathbf{s}_j, \theta) = \mathbf{v}^T_\alpha~ tanh(\mathbf{W}_\alpha[\mathbf{h}_i; \mathbf{s}_j;\theta]+\mathbf{b}_\alpha).
    \end{equation}
Here $\mathbf{v}_\alpha$, $\mathbf{W}_\alpha$, and $\mathbf{b}_\alpha$ are trainable parameters.
$f_{\alpha}(\cdot)$ measures the semantic relations between the $i$-th word in the source and the $j$-th target word to be predicted.
Such relations are also calibrated with the input's latent topic $\theta$ in order to explore and highlight topic words. 
We hence obtain the topic sensitive context vector ${\bf c}_j$ with:
\begin{equation}
   \mathbf{c}_j = \sum_{i=1}^{|\mathbf{x}|}\alpha_{ij}\mathbf{h}_i.
\end{equation}
Further, conditioned on ${\bf c}_j$, we generate the $j$-th word over the global vocabulary according to:
\begin{equation}\label{eq:copy_score}
   p_{gen} = softmax({\bf W}_{gen}[{\bf s}_j;{\bf c}_j]+{\bf b}_{gen}).
\end{equation}

In addition, we adopt copy mechanism~\cite{DBLP:conf/acl/SeeLM17} following~\citet{DBLP:conf/acl/MengZHHBC17}, which allows keywords to be directly extracted from the source input.
Specifically, we adopt a soft switcher $\lambda_j\in [0,1]$ to determine whether to copy a word from source as the $j$-th target word: 
\begin{equation}\label{eq:copy_switcher}
   \lambda_j = sigmoid(\mathbf{W}_{\lambda}[\mathbf{u}_j; \mathbf{s}_j;\mathbf{c}_j; \theta] + \mathbf{b}_{\lambda}),
\end{equation}
\noindent with $\mathbf{W}_{\lambda}$ and $\mathbf{b}_{\lambda}$ being learnable parameters. Topic information $\theta$ is also injected here to guide the switch decision.

Finally, we obtain distribution $p_j$ for predicting the $j$-th target word with the formula below:
\begin{equation}
  p_j = \lambda_j\cdot p_{gen}+ (1-\lambda_j)\cdot\sum_{i=1}^{|\mathbf{x}|}\alpha_{ij},
\end{equation}
\noindent where attention scores $\{\alpha_{ij}\}^{|\mathbf{x}|}_{i=1}$ serve as the extractive distribution over the source input.

\subsection{Jointly Learning Topics and Keyphrases}\label{ssec:joint_learn}

Our neural framework allows end-to-end learning of latent topic modeling and keyphrase generation.
We first define objective functions for the two modules respectively.

For NTM, the objective function is defined based on negative variational lower bound~\cite{DBLP:journals/corr/BleiKM16}. 
Here due to space limitation, we omit the derivation details already described in \citet{DBLP:conf/icml/MiaoGB17}, and directly give its loss function:
\begin{equation}
    \mathcal{L}_{NTM}=D_{KL}(p(\mathbf{z})\,||\,q(\mathbf{z}\,|\,\mathbf{x})) - \mathbb{E}_{q(\mathbf{z}\,|\,
    {\bf x})}[p(\mathbf{x}\,|\,\mathbf{z})],
\end{equation}
\noindent where the first term is the Kullback-Leibler divergence loss and the second term reflects the reconstruction loss.
$p({\bf z})$ denotes a standard normal prior. $q({\bf z}\,|\,{\bf x})$  and $p(\mathbf{x}\,|\,\mathbf{z})$  represent the process of BoW encoder and BoW decoder respectively.

For KG model, we minimize the cross entropy loss over all training instances:
\begin{equation}
    \mathcal{L}_{KG} = -\sum_{n=1}^N\log(Pr(\mathbf{y}_n\,|\,\mathbf{x}_n, \mathbf{\theta}_n)),
\end{equation}
where $N$ denotes the number of training instances and $\mathbf{\theta}_n$ is ${\bf x}_n$'s latent topics induced from NTM.

Finally, we define the entire framework's  training objective with the linear combination of $\mathcal{L}_{NTM}$ and $\mathcal{L}_{KG}$:
\begin{equation}\label{eq:overall_objective}
    \mathcal{L} = \mathcal{L}_{NTM} + \gamma \cdot \mathcal{L}_{KG},
\end{equation}
\noindent where the hyper-parameter $\gamma$ balances the effects of NTM and KG model.
Our two modules can be jointly trained with their parameters updated simultaneously.
For inference, we adopt beam search and generate a ranking list of output keyphrases following \citet{DBLP:conf/acl/MengZHHBC17}.

%% file: figure_input/framework.tex
\begin{figure}
\centering
\includegraphics[scale=0.5]{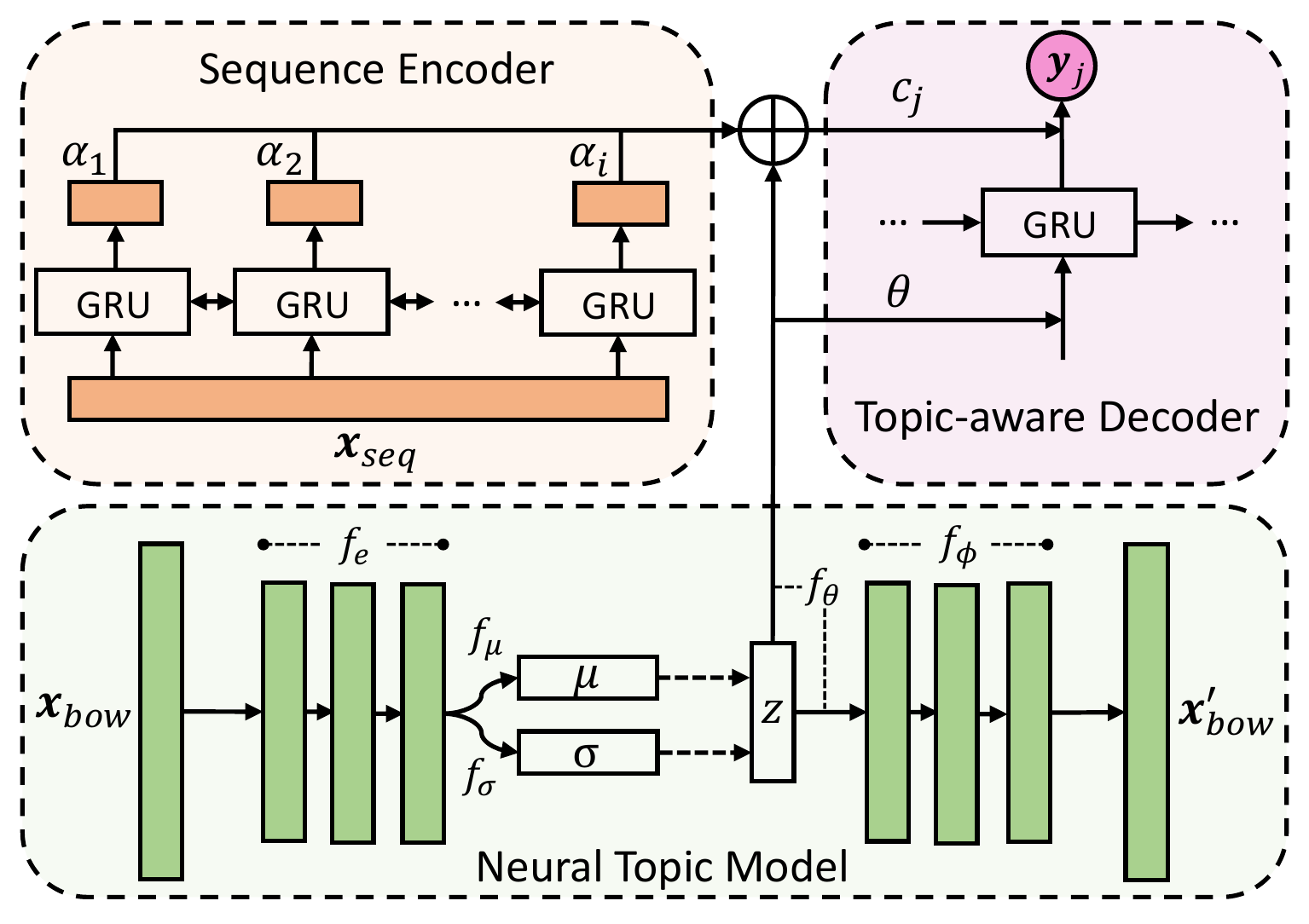}
\caption{Our topic-aware neural keyphrase generation framework (\cref{sec:model}).}\label{fig:framework}
\end{figure}

%% file: sections/exp-setup.tex
\section{Experiment Setup}

\paragraph{Datasets.} 
We conduct  experiments on three social media datasets collected from two English online platforms, \textbf{Twitter} and \textbf{StackExchange}, and a Chinese microblog website, \textbf{Weibo}.
Twitter and Weibo are microblogs encouraging users to freely post with a wide range of topics, while StackExchange, an online Q\&A forum, are mainly for question asking (with a title and a description) and seeking answers from others.
 
The Twitter dataset contains tweets from TREC 2011 microblog track.\footnote{\url{http://trec.nist.gov/data/tweets/}}
For Weibo dataset, we first tracked the real-time trending hashtags in Jan-Aug 2014,\footnote{\url{http://open.weibo.com/wiki/Trends/}} and then used them as keywords to search posts with hashtag-search API.\footnote{\url{http://www.open.weibo.com/wiki/2/}}
And the StackExchange dataset is randomly sampled from a publicly available raw corpus.\footnote{\url{https://archive.org/details/stackexchange}}

For the target keyphrases, we employ user-annotated hashtags for Twitter and Weibo following~\citet{DBLP:conf/emnlp/ZhangWGH16}, and author-assigned tags (e.g., ``\textit{artificial-intelligence}'') for StackExchange.
Posts without such keyphrase tags are hence removed from the datasets. 
Particularly, for StackExchange, we concatenate the question title together with its description as the source input. 
For Twitter and Weibo source posts, we retain tokens in hashtags (without \# symbols) for those appearing in the middle of posts, since they generally act as semantic elements and thus considered as present keyphrases~\cite{DBLP:conf/emnlp/ZhangWGH16}. 
For those appearing before or after a post, we remove the entire hashtags and regard them as absent keyphrases as is done in ~\citet{DBLP:conf/naacl/yuewang19}.

For model training and evaluation, we split the data into three subsets with $80$\%, $10$\%, and $10$\%, corresponding to training, development, and test set.
The statistics of the three datasets are shown in Table \ref{tables:dataset_stat}. 
As can be seen, over $50\%$ of the keyphrases do not appear in their source posts, thus extractive approaches will fail in dealing with these posts. 
We also observe that StackExchange exhibits different keyphrase statistics compared to either Twitter or Weibo, with more keyphrases appearing in one post and more diverse keyphrases. 

\input{tables/dataset_stat.tex}

\paragraph{Preprocessing.} For Twitter dataset, we employed Twitter preprocessing toolkit  in~\citet{DBLP:conf/semeval/BaziotisPD17a} for source post and hashtag (keyphrase) tokenization.
Chinese Weibo data was preprocessed with Jieba toolkit\footnote{\url{https://github.com/fxsjy/jieba}} for word segmentation, and English StackExchange data with natural language toolkit (NLTK) for tokenization.\footnote{\url{https://www.nltk.org/}} 

We further take the following preprocessing steps for each of the three datasets: 
First, posts with meaningless keyphrases (e.g., single-character ones) were filtered out; also removed were non-alphabetic (for English data) and retweet-only (e.g., ``\textit{RT}'') posts.
Second, links, mentions (@username), and digits were replaced with generic tags ``\textit{URL}'', ``\textit{MENT}'', and ``\textit{DIGIT}'' following~\citet{DBLP:conf/naacl/yuewang19}.
Third, a vocabulary was maintained,  with $30$K most frequent words for Twitter, and $50$K for Weibo and StackExchange each.
For BoW vocabulary of the input ${\bf x}_{bow}$ for NTM, stop words and punctuation were removed.
% following common topic model practice.

\input{tables/main_exp_full.tex}

\paragraph{Parameter Settings.}
We implement our model based on the pytorch framework in~\citet{paszke2017automatic}. For NTM, we implement it following the design\footnote{\url{https://github.com/zengjichuan/TMN}} in~\citet{DBLP:conf/emnlp/ZengLSGLK18} and set topic number $K$ to $50$. The KG model is set up mostly based on~\citet{DBLP:conf/acl/MengZHHBC17}. For its sequence encoder, we adopt two layers of bidirectional GRU and one layer of unidirectional GRU for its decoder. 
The hidden size of the GRU is $300$ (for bi-GRU, $150$ for each direction). 
For the embedding, its size is set to $150$ and values are randomly initialized. 
We apply Adam~\cite{DBLP:conf/iclr/KingmaB14} with initial learning rate as $1e-3$. 
In training process, gradient clipping $=1.0$ is conducted to stabilize the training. 
Early-stopping strategy~\cite{caruana2001overfitting} is adopted based on the validation loss. 
Before joint training, we pretrain NTM for $100$ epochs and KG model for $1$ epoch as the convergence speed of NTM is much slower than the KG model.
We empirically set the $\gamma=1.0$ for balancing NTM and KG loss (Eq.~\ref{eq:overall_objective}) and iteratively update the parameters in each module and then their combination in turn.

\paragraph{Comparisons.}

In comparison, we first consider a simple baseline selecting majority keyphrases (henceforth \textsc{Majority}) --- the top $K$ keyphrases ranked by their frequency in training data are used as the keyphrases for all test instances.
We also compare with the following extractive baselines, where n-grams ($n=1,2,3$) in source posts are ranked by TF-IDF scores (henceforth \textsc{TF-IDF}),  TextRank algorighm~\cite{mihalcea2004textrank} (henceforth \textsc{TextRank}), and KEA system~\cite{DBLP:conf/dl/WittenPFGN99} (henceforth \textsc{KEA}). We also compare with a neural state-of-the-art keyphrase extraction model based on sequence tagging~\cite{DBLP:conf/emnlp/ZhangWGH16} (henceforth \textsc{Seq-Tag}).
In addition, we take the following state-of-the-art keyphrase generation models into consideration: seq2seq model with copy mechanism~\cite{DBLP:conf/acl/MengZHHBC17} (henceforth \textsc{Seq2Seq-Copy}) and its variation \textsc{Seq2Seq} without copy mechanism,  \textsc{Seq2Seq-Corr}~\cite{DBLP:conf/emnlp/ChenZ0YL18} exploiting keyphrase correlations, and  \textsc{TG-Net}~\cite{DBLP:conf/aaai/wchen19} jointly modeling of titles and descriptions (thereby only tested on StackExchange).

%% file: tables/dataset_stat.tex
\begin{table}
\begin{center}
\resizebox{0.49\textwidth}{!}{
\begin{tabular}{|l|rrrr|}
\hline
\multirow{2}{*}{\textbf{Source posts}}& \# of  & Avg len & \# of KP & Source\\ & posts & per post & per post & vocab\\
\hline
Twitter &44,113 & 19.52 &1.13 &34,010\\
Weibo& 46,296&33.07&1.06&98,310\\
StackExchange&49,447&87.94&2.43&99,775\\
\hline
\hline
\multirow{2}{*}{\textbf{Target KP}}& \multirow{2}{*}{$|$KP$|$} & Avg len & \% of & Target\\
&& per KP&  abs KP&vocab\\
\hline
Twitter&4,347&1.92&71.35&4,171\\
Weibo&2,136&2.55&75.74&2,833\\
StackExchange&12,114&1.41&54.32&10,852\\
\hline
\end{tabular}
}
\end{center}

\caption{Data statistics of source posts (on the top) and target keyphrases (on the bottom). Avg len: the average number of tokens. KP: keyphrases. Abs KP: absent keyphrases. $|$KP$|$: the number of distinct keyphrases.
} \label{tables:dataset_stat}

\end{table}

%% file: tables/main_exp_full.tex
\begin{table*}[ht]
\begin{center}
\resizebox{0.99\textwidth}{!}{
\begin{tabular}{|l|ccc|ccc|ccc|}
\hline
\multirow{2}{*}{\textbf{Model}}&\multicolumn{3}{c|}{\textbf{Twitter}}&\multicolumn{3}{c|}{\textbf{Weibo}}&\multicolumn{3}{c|}{\textbf{StackExchange}}\\
\cline{2-10}

&F1@1 &F1@3 &MAP 
&F1@1 &F1@3 &MAP
&F1@3 &F1@5 &MAP\\
\cline{2-10}
\hline
\underline{\textbf{Baselines}}&&&&&&&&&\\

\textsc{Majority} 
&9.36 &11.85 & 15.22 
&4.16 &3.31 &5.47
&1.79 &1.89 & 1.59 \\

\textsc{TF-IDF} 
&1.16 &1.14 &1.89 
&1.90 &1.51 &2.46
&13.50 &12.74 &12.61\\

\textsc{TextRank} 
&1.73 &1.94 & 1.89
&0.18 &0.49 &0.57
&6.03 &8.28  &4.76\\

\textsc{KEA} 
&0.50 &0.56 &0.50
&0.20 &0.20 &0.20
&15.80 &15.23 &14.25\\

\hline
\hline

\underline{\textbf{State of the arts}}&&&&&&&&&\\

\textsc{Seq-Tag} 
&22.79\scriptsize{$\pm$0.3} &12.27\scriptsize{$\pm$0.2} &22.44\scriptsize{$\pm$0.3} 
&16.34\scriptsize{$\pm$0.2} &8.99\scriptsize{$\pm$0.1} &16.53\scriptsize{$\pm$0.3}
&17.58\scriptsize{$\pm$1.6} &12.82\scriptsize{$\pm$1.2} &19.03\scriptsize{$\pm$1.3}\\

\textsc{Seq2Seq}
&34.10\scriptsize{$\pm$0.5} &26.01\scriptsize{$\pm$0.3} &41.11\scriptsize{$\pm$0.3}
&28.17\scriptsize{$\pm$1.7}&20.59\scriptsize{$\pm$0.9}&34.19\scriptsize{$\pm$1.7}
&22.99\scriptsize{$\pm$0.3} &20.65\scriptsize{$\pm$0.2} &23.95\scriptsize{$\pm$0.3}\\

\textsc{Seq2Seq-Copy}
&\underline{36.60}\scriptsize{$\pm$1.1} &\underline{26.79}\scriptsize{$\pm$0.5} &\underline{43.12}\scriptsize{$\pm$1.2} 
&\underline{32.01}\scriptsize{$\pm$0.3} &\underline{22.69}\scriptsize{$\pm$0.2} &\underline{38.01}\scriptsize{$\pm$0.1}
&31.53\scriptsize{$\pm$0.1}  &27.41\scriptsize{$\pm$0.2} &33.45\scriptsize{$\pm$0.1}\\

\textsc{Seq2Seq-Corr}
&34.97\scriptsize{$\pm$0.8} &26.13\scriptsize{$\pm$0.4} &41.64\scriptsize{$\pm$0.5}
&31.64\scriptsize{$\pm$0.7} &22.24\scriptsize{$\pm$0.5} &37.47\scriptsize{$\pm$0.8}
&30.89\scriptsize{$\pm$0.3} &26.97\scriptsize{$\pm$0.2} &32.87\scriptsize{$\pm$0.6}\\

\textsc{TG-Net}
&- &- &-
&- &- &-
&\underline{32.02}\scriptsize{$\pm$0.3} &\underline{27.84}\scriptsize{$\pm$0.3} &\underline{34.05}\scriptsize{$\pm$0.4}\\

\hline
\hline

Our model
&\textbf{38.49}\scriptsize{$\pm$0.3} &\textbf{27.84}\scriptsize{$\pm$0.0} &\textbf{45.12}\scriptsize{$\pm$0.2}
&\textbf{34.99}\scriptsize{$\pm$0.3} &\textbf{24.42}\scriptsize{$\pm$0.2} &\textbf{41.29}\scriptsize{$\pm$0.4}
&\textbf{33.41}\scriptsize{$\pm$0.2} &\textbf{29.16}\scriptsize{$\pm$0.1} &\textbf{35.52}\scriptsize{$\pm$0.1}\\

\hline
\end{tabular}
}
\end{center}
\caption{Main comparison results displayed with average scores (in \%) and their standard deviations over the results with $5$ sets of random initialization seeds. Boldface scores in each column indicate the best results. Our model significantly outperforms all comparisons on all three datasets ($p<0.05$, paired t-test).
}\label{tables:main_exp_full}
\vspace{-0.5em}
\end{table*}

%% file: sections/exp-result.tex
\section{Experimental Results}

In the experiment, we first evaluate our performance on keyphrase prediction (\cref{ssec:main_exp}). 
Then, we study whether jointly learning keyphrase generation can in turn help produce coherent topics (\cref{ssec:latent_topic}). 
At last, further discussions (\cref{ssec:discussion}) are presented with an ablation study, a case study, and an analysis for varying text genres.

\subsection{Keyphrase Prediction Results}\label{ssec:main_exp}

In this section, we examine our performance in predicting keyphrases for social media.
We first discuss the main comparison results, followed by a discussion for present and absent keyphrases.

Popular information retrieval metrics macro-average F1@K and mean average precision (MAP) are adopted for evaluation.
Here for Twitter and Weibo, most posts are tagged with one keyphrase on average (Table~\ref{tables:dataset_stat}), thus F1@1 and F1@3 are reported.
For StackExchange, we report F1@3 and F1@5, because on average, posts have $2.4$ keyphrases.
MAP is measured over the top $5$ predictions for all three datasets. 
For keyphrase matching, we consider keyphases after stemmed by Porter Stemmer following~\citet{DBLP:conf/acl/MengZHHBC17}.

\paragraph{Main Comparison Discussion.}

Table~\ref{tables:main_exp_full} shows the main comparison results on our three datasets, where higher scores indicate better performance.
From all three datasets, we observe:

$\bullet$~\textbf{\textit{Social media keyphrase prediction is challenging.}}
As can be seen, all simple baselines give poor performance. 
This indicates that predicting keyphrases for social media language is a challenging task.
It is impossible to rely on simple statistics or rules to yield good results.

$\bullet$~\textbf{\textit{Seq2seq-based keyphrase generation models are effective.} }
Compared to the extractive baselines and \textsc{Seq-Tag},
seq2seq-based models perform much better. 
It is because social media's informal language style results in a large amount of absent keyphrases (Table~\ref{tables:dataset_stat}), which is impossible for extractive methods to make correct predictions. 
We also find \textsc{Seq2seq-copy} better than \textsc{Seq2seq}, suggesting the effectiveness to combine source word extraction with word generation when predicting keyphrases.

$\bullet$~\textbf{\textit{Latent topics are consistently helpful for indicating keyphrases.}}
It is observed that our model achieves the best results, significantly outperforming all comparisons by a large margin.
This shows the usefulness of leveraging latent topics in keyphrase prediction.
Interestingly, compared with StackExchange, we achieve larger improvements for Twitter and Weibo, both exhibiting more informal nature and prominent word order misuse.
For such text genres, latent topics, learned under BoW assumption, are more helpful.

Also, the following interesting points can be observed by comparing results across datasets:

$\bullet$~\textbf{\textit{Keyphrase generation is more challenging for StackExchange.}}
When MAP scores of seq2seq-based methods are compared over the three datasets, we find that the scores on StackExchange are generally lower.
It is probably attributed to the data characteristics of more diverse keyphrases and larger target vocabulary (Table~\ref{tables:dataset_stat}).

$\bullet$~\textbf{\textit{Twitter and Weibo data is noisier.}}
We notice that \textsc{TF-IDF}, \textsc{TextRank}, and \textsc{KEA} perform much worse than \textsc{Majority}, while the opposite is observed on StackExchange.
It is because Twitter and Weibo, as microblogs, contain shorter posts (Table~\ref{tables:dataset_stat}) and exhibit more informal language styles.
In general, models relying on simple word statistics would suffer from such noisy data.

\input{figure_input/present_absent.tex}

\paragraph{Present and Absent Keyphrase Prediction.}

We further discuss how our model performs in producing present and absent keyphrases. 
The comparison results with all neural-based models are shown in Figure~\ref{fig:present_absent}.
Here F1@1 is adopted for evaluating the prediction of present keyphrases and recall@5 for absent keyphrases. 

The results indicate that our model consistently outperforms comparison models in predicting either absent or present keyphrases.
Also, interestingly, copy mechanism seems to somehow sacrifice the performance on absent keyphrase generation for correctly extracting the present ones. 
Such side effects, however, are not observed on our model.
It is probably attributed to our ability to associate posts with corpus-level topics, hence enabling absent keywords from other posts to be ``copied''.
This observation also demonstrates the latent topics can help our model to better decide whether to copy (Eq.~\ref{eq:copy_switcher}).

\subsection{Latent Topic Analysis}\label{ssec:latent_topic}

We have shown latent topics useful for social media keyphrase generation in \cref{ssec:main_exp}. 
Here we analyze whether our model can learn meaningful topics.

\paragraph{Coherence Score Comparison.}
We first evaluate topic coherence with an automatic $C_V$ measure. 
Here we employ Palmetto toolkit\footnote{\url{https://github.com/dice-group/Palmetto/}}~\cite{DBLP:conf/wsdm/RoderBH15} on the top $10$ words from each latent topic following~\citet{DBLP:conf/emnlp/ZengLSGLK18}. 
The results are only reported on English Twitter and StackExchange because Palmetto does not support Chinese.
For comparisons, we consider LDA (implemented with a gensim LdaMulticore package\footnote{\url{https://pypi.org/project/gensim/}}), BTM\footnote{\url{https://github.com/xiaohuiyan/BTM}}~\cite{yan2013biterm} (a state-of-the-art topic model specifically for short texts), and NTM~\cite{DBLP:conf/icml/MiaoGB17}. 
For LDA and BTM, we run Gibbs sampling with $1,000$ iterations  to ensure convergence.
From the results in Table~\ref{tables:topic_coherence}, we observe that our model outperforms all the comparison topic models by large margins, which implies that jointly exploring keyphrase generation can in turn help produce coherent topics.

\input{tables/topic_coherence.tex}

\paragraph{Sample Topics.}
To further evaluate whether our model can produce coherent topics qualitatively, we probe into some sample words (Table~\ref{tables:topic_sample}) reflecting the topic ``\textit{super bowl}'' discovered by various models from Twitter.
As can be seen, there are mixed non-topic words
\footnote{Non-topic words refer to words that cannot clearly indicate the corresponding topic, including off-topic words more likely to reflect other topics.}
in LDA's, BTM's, and NTM's sample topic.
Compared with them, our inferred topic looks more coherent.
For example, ``\textit{steeler}'' and ``\textit{packer}'', names of \textit{super bowl} teams, are correctly included into the cluster.

\input{tables/topic_sample.tex}

\subsection{Further Discussions}\label{ssec:discussion}
% We have shown our superiority in keyphrase prediction (\cref{ssec:main_exp}) and latent topic induction (\cref{ssec:latent_topic}). 
% To analyze the underlying reasons, below we further discuss our model in depth.

\paragraph{Ablation Study.}

We compare the results of our full model and its four ablated variants to analyze the relative contributions of topics on different components.
The results in Table~\ref{tabs:ablation} indicate the competitive effect of topics on decoder attention and that on hidden states, but combining them both help our full model achieve the best performance.
We also observe that pre-trained topics only bring a small boost, indicated by the close scores yielded by our model (\textit{separate train}) and \textsc{Seq2Seq-Copy}.
This suggests that the joint training is crucial to better absorb latent topics. 

\input{tables/ablation.tex}

\paragraph{Case Study.}

We feed the tweet $S$ in Table~\ref{tables:intro-example} into both \textsc{Seq2Seq-copy} and our model. 
Eventually our model correctly predicts the keyphrase as ``\textit{super bowl}'' while \textsc{Seq2Seq-copy} gives a wrong prediction ``\textit{team follow back}'' (posted to ask other to follow back).
To analyze the reason behind, we visualize the attention weights of two models in Figure~\ref{fig:attn_vis}.
It can be seen that both models highlight the common word ``\textit{team}'', which frequently appears in ``\textit{team follow back}''-tagged tweets. 
By joint modeling of latent topics, our model additionally emphasizes topic words ``\textit{yellow}'' and ``\textit{pants}'', which are signals indicating a super bowl team \textit{steeler} (also reflected in the $1^{st}$ topic) and thus helpful to correctly generate ``\textit{super bowl}'' as its keyphrase.
Without such topic guidance, \textsc{Seq2seq-copy} wrongly predicts a common but unrelated term ``\textit{team follow back}''.

\input{figure_input/attn_vis.tex}

\paragraph{Topic-Aware KG for Other Text Genres.}

We have shown the effectiveness of latent topics on social media keyphrase generation.
To examine how they affect in identifying keyphrases for well-edited language, we also experiment on the traditional scientific article datasets~\cite{DBLP:conf/acl/MengZHHBC17}, but limited improvements are observed. 
Latent topics can better help keyphrase generation on social media, probably because there are larger proportion of keyphrases with absent words (Figure~\ref{fig:absent_comp}), where latent topics can cluster relevant posts and enrich the source contexts.
Another possible reason lies in that social media language exhibits prominent arbitrary word orders. 
Thus latent topics, learned under BoW assumption, can better provide useful auxiliary features.

\input{figure_input/absent_comp.tex}

%% file: figure_input/present_absent.tex
\begin{figure}[H]
\centering
\includegraphics[scale=0.45, width=7.8cm, height=5.5cm, trim=0 0 0 0]{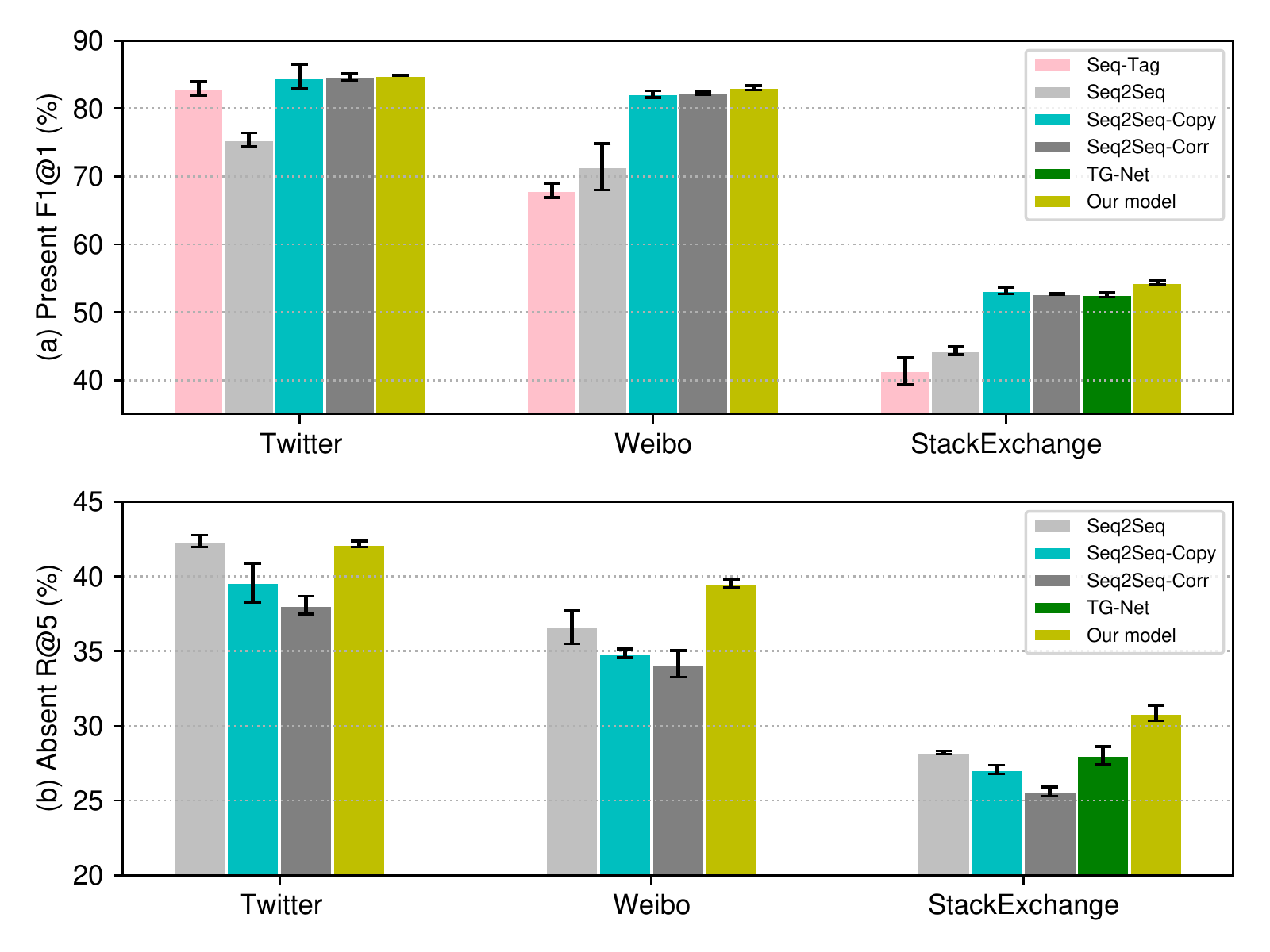}
% \vspace{-1em}
\caption{The prediction results for present (on the top) and absent keyphrases (on the bottom, R@5: recall@5). 
For present cases, from left to right shows the results of \textsc{Seq-Tag}, \textsc{Seq2Seq}, \textsc{Seq2Seq-Copy}, \textsc{Seq2Seq-Corr}, \textsc{TG-Net} (only for StackExchange), and our model. For absent cases, models (except  \textsc{Seq-Tag}) are shown in the same order. }\label{fig:present_absent}
\end{figure}

%% file: tables/topic_coherence.tex
\begin{table}[t]
\centering
\resizebox{0.38\textwidth}{!}{
\begin{tabular}{|l|cc|}
\hline
\textbf{Datasets} & \textbf{Twitte}r &\textbf{StackExchange} \\
\hline
LDA & 41.12 & 35.13\\
BTM & 43.12 & 43.52\\
NTM & 43.82 & 43.04\\
 Our model &\textbf{46.28} &\textbf{45.12}\\
\hline
\end{tabular}
}
\caption{$C_V$ topic coherence score comparison on our two English datasets. Higher scores indicate better coherence. Our model produces the best scores.}
\label{tables:topic_coherence}

\end{table}

%% file: tables/topic_sample.tex
\begin{table}[H]
\centering
\resizebox{0.48\textwidth}{!}{  
\begin{tabular}{|p{1.2cm}|p{6cm}|}

\hline

\multirow{2}{*}{LDA}& 
bowl super \textcolor{red}{\uline{quote}} steeler \textcolor{red}{\uline{jan}} watching \textcolor{red}{\uline{egypt}} playing glee \textcolor{red}{\uline{girl}} \\
\hline
\multirow{2}{*}{BTM}& 
bowl super anthem national christina aguilera fail \textcolor{red}{\uline{word}} brand playing \\

\hline
\multirow{2}{*}{NTM} &
super bowl eye \textcolor{red}{\uline{protester}} winning watch halftime  ship sport \textcolor{red}{\uline{mena}} \\

\hline
Our model&  
bowl super yellow green packer steeler nom commercial win winner\\
\hline
\end{tabular}
}
\caption{Top 10 terms for latent topics ``\textit{super bowl}''. Red and underlined words indicate \textcolor{red}{\uline{non-topic words}}.}
\label{tables:topic_sample}

\end{table}

%% file: tables/ablation.tex
\begin{table}[H]
\centering
\resizebox{0.49\textwidth}{!}{
\begin{tabular}{|l|rrr|}
\hline
\textbf{Model} &\textbf{Twitter} & \textbf{Weibo} & \textbf{SE}\\
\hline
\textsc{Seq2Seq-Copy}
&36.60 &32.01 &31.53 \\

Our model (\textit{separate train})   
&36.75  &32.75 &31.78   \\ %34.75

Our model (\textit{w/o topic-attn})
&37.24 &32.42 &32.34\\

Our model (\textit{w/o topic-state})
&37.44 &33.48 &31.98 \\ %35.44

\hline
Our full model   
&\textbf{38.49} &\textbf{34.99} &\textbf{33.41}\\

\hline
\end{tabular}
}

\caption{Comparison results of our ablation models on three datasets (SE: StackExchange) ---
\textit{separate train}: our model with pre-trained latent topics;
\textit{w/o topic-attn}: decoder attention without topics (Eq.~\ref{eq:attn_score});
\textit{w/o topic-state}: decoder hidden states without topics (Eq.~\ref{eq:dec_input}).
We report F1@1 for Twitter and Weibo, F1@3 for StackExchange. Best results are in bold. 
}\label{tabs:ablation}

\end{table}

%% file: figure_input/attn_vis.tex
\begin{figure}[H]
\centering
\includegraphics[scale=0.9, trim={0 0 0 5}]{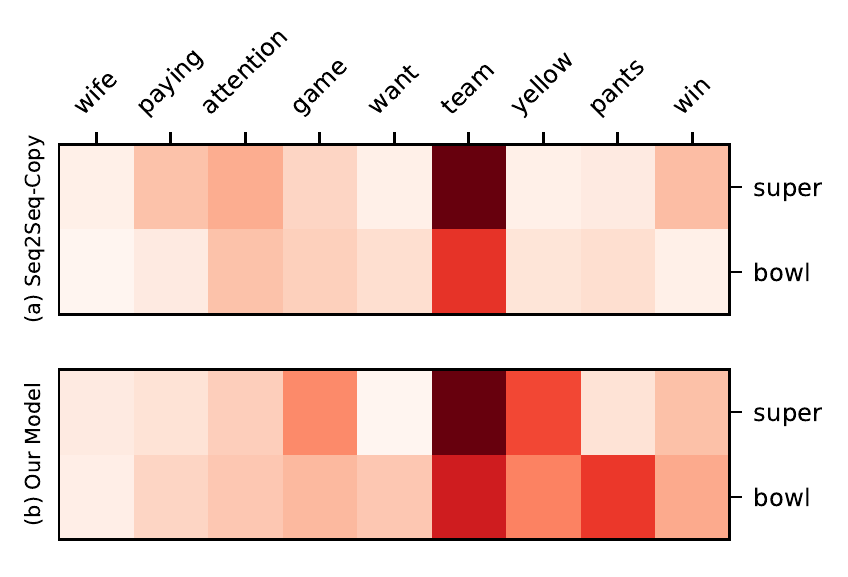}

\begin{tabular}{|l|l|}

\hline
$1^{st}$ Topic & steeler national team packer win\\
\hline
\end{tabular}

\caption{Attention visualization for the sample post in Table~\ref{tables:intro-example}. Only non-stopwords are selected. The table below shows the top five words for the $1^{st}$ topic.}\label{fig:attn_vis}
\end{figure}

%% file: figure_input/absent_comp.tex
\begin{figure}[H]
\centering
\includegraphics[scale=0.49, trim={0 0 0 0}]{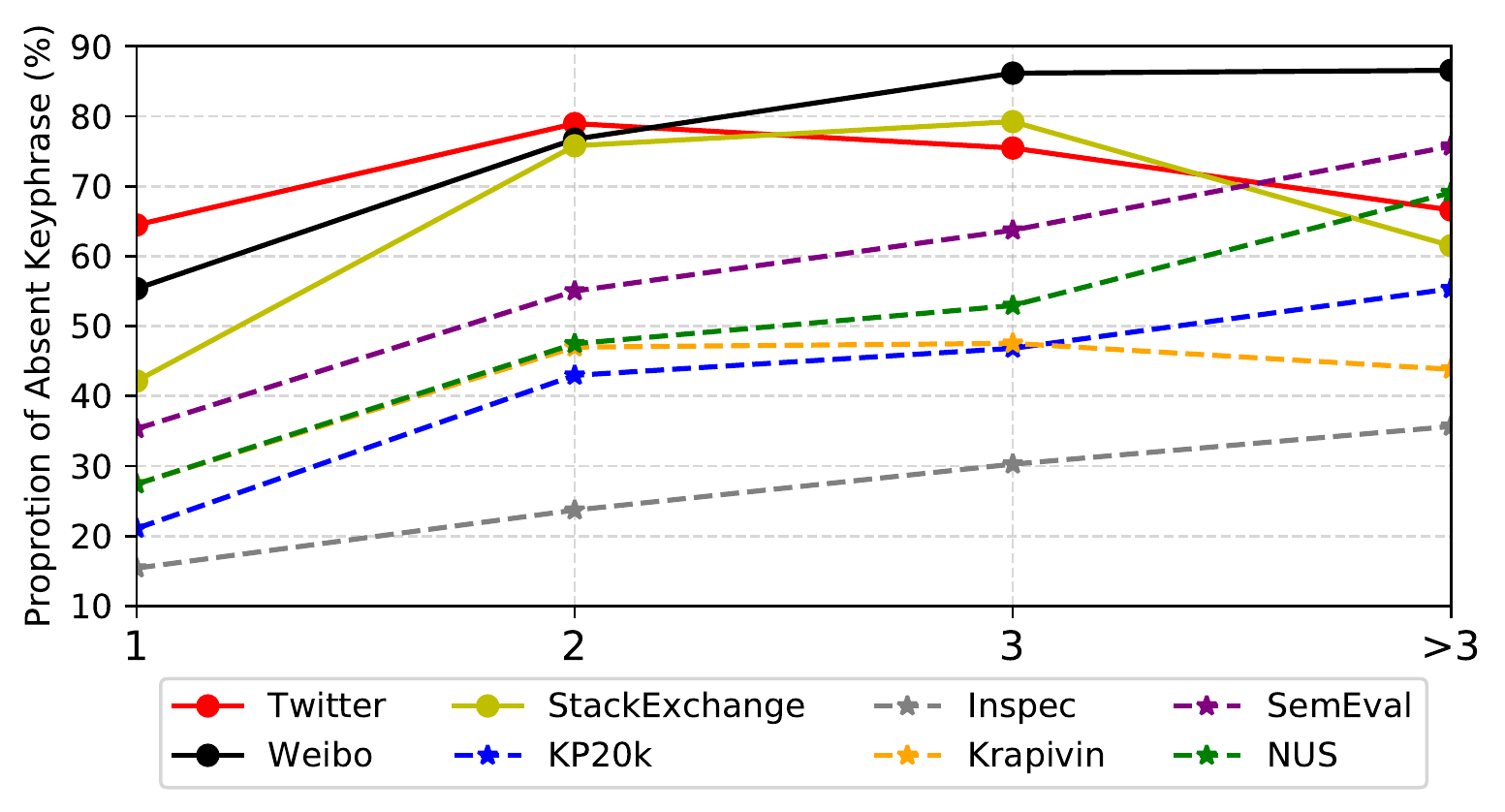}

\caption{Proportion of absent $n$-gram keyphrases ($n$: $1,2,3,>3$). The dashed lines with `*' marks: the five scientific article datasets used in~\citet{DBLP:conf/acl/MengZHHBC17}.}\label{fig:absent_comp}
\end{figure}

%% file: sections/conclusion.tex
\section{Conclusion and Future Work}
We have presented a novel social media keyphrase generation model that allows the joint learning of latent topic representations. 
Experimental results on three newly constructed social media datasets show that our model significantly outperforms state-of-the-art methods in keyphrase prediction, meanwhile produces more coherent topics. 
Further analysis interprets our superiority to discover key information from noisy social media data.

In the future, we will explore how to explicitly leverage the topic-word distribution to further improve the performance. 
Also, our topic-aware neural keyphrase generation model can be investigated in a broader range of text generation tasks.

%% file: sections/ack.tex
\section*{Acknowledgements}
This work is supported by the Research Grants Council of the
Hong Kong Special Administrative Region, China (No. CUHK 14208815 and No. CUHK 14210717 of the General Research Fund). We thank ACL reviewers for their insightful suggestions on various aspects of this work.